\pdfoutput=1

\documentclass[11pt]{article}

\usepackage[final]{acl}

\usepackage{times}
\usepackage{latexsym}

\usepackage[dvipsnames]{xcolor}
\usepackage[T1]{fontenc}

\usepackage[utf8]{inputenc}

\usepackage{microtype}

\usepackage{inconsolata}

\usepackage{graphicx}
\usepackage{cuted}
\usepackage{color,soul}
\usepackage{amsfonts}
\usepackage{amsmath}
\usepackage{multirow}
\usepackage{float}

\title{Retrieval Augmented Generation based context discovery for ASR}

\author{
  \textbf{Dimitrios Siskos\textsuperscript{1}},
  \textbf{Stavros Papadopoulos\textsuperscript{1}},
  \textbf{Pablo Peso Parada\textsuperscript{2}},
  \\
  \textbf{Jisi Zhang\textsuperscript{2}},
  \textbf{Karthikeyan Saravanan\textsuperscript{2}},
  \textbf{Anastasios Drosou\textsuperscript{1}}
\\
\\
  \textsuperscript{1}Information Technologies Institute, Center for Research and Technology Hellas,\\ Thessaloniki, Greece \\
  \textsuperscript{2}Samsung Electronics R\&D Institute UK (SRUK), London, United Kingdom
\\
  \small{
    \href{mailto:d.siskos@iti.gr}{d.siskos@iti.gr},
    \href{mailto:spap@iti.gr}{spap@iti.gr},
    \href{mailto:p.parada@samsung.com}{p.parada@samsung.com},
    \href{mailto:jisi.zhang@samsung.com}{jisi.zhang@samsung.com},
    \href{mailto:k1.saravanan@samsung.com}{k1.saravanan@samsung.com}
    \href{mailto:drosou@iti.gr}{drosou@iti.gr},
  }
}

\begin{document}
    \maketitle
    \begin{abstract}
    This work investigates retrieval augmented generation as an efficient strategy for automatic context discovery in context-aware Automatic Speech Recognition (ASR) system, in order to improve transcription accuracy in the presence of rare or out-of-vocabulary terms. However, identifying the right context automatically remains an open challenge. This work proposes an efficient embedding-based retrieval approach for automatic context discovery in ASR. To contextualize its effectiveness, two alternatives based on large language models (LLMs) are also evaluated: (1) large language model (LLM)-based context generation via prompting, and (2) post-recognition transcript correction using LLMs. Experiments on the TED-LIUMv3, Earnings21 and SPGISpeech demonstrate that the proposed approach reduces WER by up to 17\% (percentage difference) relative to using no-context, while the oracle context results in a reduction of up to 24.1\%.
    \end{abstract}

    \section{Introduction}
    
    \begin{figure*}[t]
      \includegraphics[width=1\linewidth]{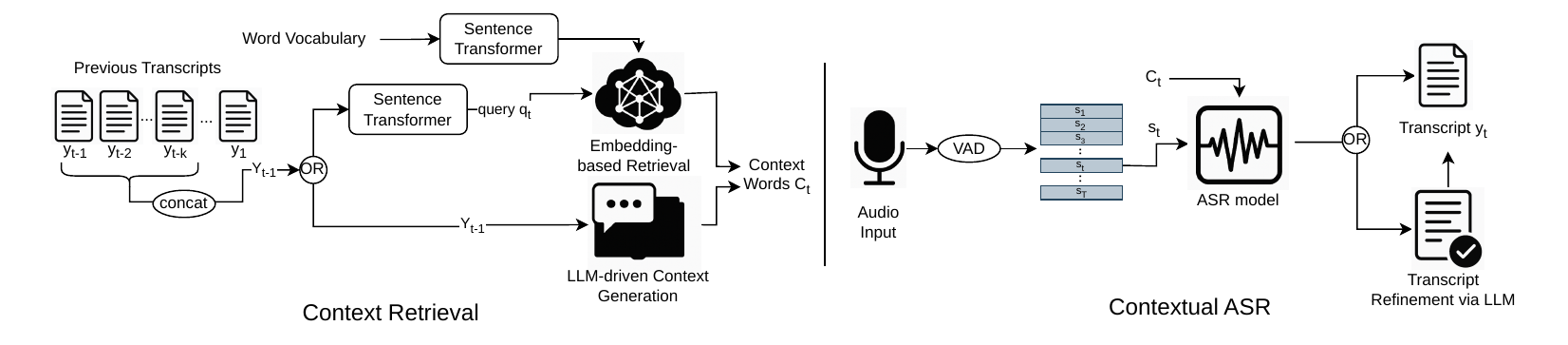} 
      \caption {Overview of the proposed context-aware ASR pipeline. Context for each segment is extracted using either an embedding-based retrieval method or LLM-based prompt generation, both conditioned on the preceding $k$ segment captions. Audio is segmented via Voice Activity Detection (VAD). The selected context is provided to a contextual ASR system. Optionally, a post-ASR LLM correction module refines the transcript. The final output is the concatenation of all the individual segment transcripts.}
      \label{fig:pipeline}
    \end{figure*}
    Automatic Speech Recognition (ASR) systems have gained considerable development over the last few years and provide high transcription accuracy over an extensive spectrum of tasks and benchmarks \cite{Park2019SpecAugment, Gulati2020Conformer, Baevski2020wav2vec}. Even though ASR systems have grown tremendously, ASR systems' performance continues to degrade under conditions of low-occurrence or out-of-vocabulary words like names, specialized domain names, and user-personal references \cite{Jain2020Contextual, Fu2023Robust}. 
    To address this, several strategies have emerged, including ASR personalization \cite{Gourav2021, MysoreSathyendra2022}, text injection \cite{Sainath2023, Wu2024}, and contextual biasing (CB) \cite{huang2024improving, meng2024text}. Among these, CB has received significant attention due to its effectiveness in improving recognition of rare and user-specific terms.
    
    One of the core challenges of contextual ASR is the definition and extraction of context. The success of biasing depends not only on model architecture but also on how well contextual material aligns to audio semantics \cite{Han2022FineCoS}. 
        
    Several works have explored CB by tightly integrating with the ASR model through adapters or attention mechanisms over entity catalogs. Muralidhar et al. \cite{muralidhar2023retrieve} propose a retrieval-based personalization method, where ASR encoder representations query an infrequent entity catalog to retrieve phonetically similar candidates via a contextual adapter. Tong et al. \cite{tong2023slot} extend this by using slot-specific catalogs and combining entity embeddings with ASR outputs through a cross-attention mechanism. These methods generally rely on fine-tuning the ASR backbone and assume access to structured entity catalogs, often annotated by domain/slot type.
    
    Recent research has explored large language models (LLMs) for contextual ASR tasks. Xiao et al. \cite{xiao2025contextual} build a contextual knowledge base from custom vocabularies and documents, using a first-pass ASR output to prompt an LLM for error correction. Sun et al. \cite{sun2024contextual} fine-tune an LLM for token prediction and entity classification, applying it to second-pass hypothesis rescoring. These approaches, while effective, typically involve fine-tuning and curated entity prompts, which can limit adaptability across domains or resource-constrained settings.
    
    Finally, Mathur et al. \cite{mathur-etal-2024-doc} propose DOC-RAG, a domain-sensitive framework that builds domain-specific corpora and uses co-occurrence matrices to estimate next-word probabilities for ASR rescoring. Retrieval-based approaches such as those in \cite{muralidhar2023retrieve, tong2023slot} also leverage similarity search mechanisms, but their retrieval is conditioned on either learned audio embeddings or manually annotated slot labels, making them dependent on supervised signals and ASR model internals.
    
    The majority of existing methods assume structured catalogs \cite{muralidhar2023retrieve, tong2023slot}, require architectural modifications to ASR models \cite{tong2023slot, tong2023hierarchical}, or depend on domain-specific supervision \cite{sun2024contextual, mathur-etal-2024-doc}, limiting their flexibility in general or black-box settings. Inspired by prior work, this study proposes a lightweight, modular approach that avoids such constraints. Specifically, it proposes an embedding-based retrieval strategy using pre-trained MiniLM embeddings over a large vocabulary, entirely decoupled from the ASR model architecture.  This method is evaluated alongside two LLM-based alternatives: a prompt-based context generator and a post-ASR transcript corrector.
    
    Earlier context-biasing techniques are deeply integrated into the ASR model itself - requiring custom layers, retraining, or access to model internals - so can't be run on the black-box recognizer used in this work. Consequently, we benchmark only “plug-and-play” methods in a common pipeline that mirrors real-world deployment scenarios (identical audio inputs, metrics, and latency limits).
    
    This paper has three main contributions:
    \begin{itemize}
        \item First, it introduces a novel, model-agnostic, retrieval-based context construction technique using MiniLM embeddings on top of a frozen vocabulary to enable efficient context generation. 
        \item Second, it offers an experimental comparison of plug-and-play solutions to automatic context combination in ASR, including LLM-based context generation and post-hoc transcript fine-tuning. 
        \item Third, it provides an end-to-end evaluation of all methods in a shared pipeline that works without fine-tuning the ASR model, using black-box models and real-world long-form audio corpora.
    \end{itemize}

    The rest of the paper is organized as follows: Section \ref{sec:methods} presents the proposed methodology, Sections \ref{sec:setup} and \ref{sec:results} the experimental setup and results, and finally the paper concludes in Section \ref{sec:conlcusion}.
    
    \section{Methods} \label{sec:methods}
    
    An overview of the proposed approach is presented in Figure \ref{fig:pipeline}, and explained in the next subsections.
    
    \subsection{Contextual ASR}
    The ASR model is defined as a conditional decoder:
    $\hat{y}_t = \mathcal{A}(s_t \mid \text{C}_t)$
    where $\hat{y}_t$ is the predicted transcript of segment $s_t$, conditioned on the context list $\text{C}_t$.
    
    As a further use of LLMs, their use for post-ASR corrections has been investigated. The correction model $\mathcal{M}_{\text{fix}}$ produces a revised transcript, which is the original sentence with corrected typos and misspellings: $\hat{y}_t^{\text{corr}} = \mathcal{M}_{\text{fix}}(\hat{y}_t, C_t, H_{t-1}, P_{fix})$ where $H_{t-1}$ denotes the corrected transcript history up to segment $y_{t-1}$ and $P_{fix}$ the prompt used to instruct the LLM to fix the transcript. The use of $\hat{y}_t^{\text{corr}}$ for context is hereby denoted as \verb|LLM-fix|. Llama3.2 (3B) is utilized for all experiments requiring an LLM without loss of generality.
    
    \subsection{Context Construction}
    Let $\hat{y_t}$ be the transcript generated by ASR for segment $s_t$ (identified by a Voice Activity Detection (VAD) module) and let the transcript of the previous $k$ segments be $\hat{Y_{t-1}} = concat(\hat{y_{t-1}},...,\hat{y_{t-k}})$. $\hat{Y_{t-1}}$ is utilized to retrieve words contextually relevant to $s_t$ which are merged into a context list that's passed to the ASR system for processing.
    
    Each word $w_i$ in the vocabulary $\mathcal{V} = \{w_1, \ldots, w_{|\mathcal{V}|}\}$ is embedded using a function $f: \mathcal{V} \rightarrow \mathbb{R}^d$, such that $f(w_i) = \mathbf{v}_i \in \mathbb{R}^d$. The MiniLM (all-MiniLM-L6-v2) is utilized, similarly to BertTopic\footnote{https://maartengr.github.io/BERTopic}, where it is employed to encode textual segments and identify semantically related terms within a topic. The query vector $\mathbf{q}_t \in \mathbb{R}^d$ for segment $s_t$ is computed as: $\mathbf{q}_t = f(\hat{Y_{t-1}})$, effectively retrieving terms that align with the latent "spoken topic" of the segment, even when they are not explicitly mentioned.
    
    Top-$N$ context words are selected by maximizing cosine similarity:
    
    $$
    \text{C}_t^{\text{rag}} = \text{TopN}_{w_i \in \mathcal{V}} \left( \cos(\mathbf{q}_t, \mathbf{v}_i) \right)
    $$
    
    Efficient nearest-neighbor search is performed via FAISS\cite{douze2024faiss} indexing. The use of $\text{C}_t^{\text{rag}}$ for context is hereby denoted as \verb|CB-RAG|.
    
    Alternatively, an LLM-based approach for context construction is examined. Given the same prior window $\hat{Y_{t-1}}$ and a prompt $P_{gen}$ are passed to an LLM $\mathcal{M}_{\text{gen}}$, returning:
    
    $$
    \text{C}_t^{\text{llm}} = \mathcal{M}_{\text{gen}}(\hat{Y_{t-1}},P_{gen})
    $$
    
    The output context is post-processed to remove duplicates and stopwords. The use of $\text{C}_t^{\text{llm}}$ for context is hereby denoted as \verb|CB-LLM|.
        
    \subsection{Baselines}
    In order to measure the impact of context information quantitatively, two baseline reference strategies are utilized: the lower and upper bound of performance. 
    Let the ground-truth transcript for segment $s_t$ be denoted as $y_t$. The Oracle context (lower bound on WER) is constructed as: $\text{C}_t^{\text{oracle}} = \{ w \in y_t \mid w \notin \mathcal{S} \}$
    where $\mathcal{S}$ denotes a predefined stopword list\footnote{NLTK Corpus Stopwords https://www.nltk.org/nltk\_data/}. This assumes perfect knowledge of all the context terms and corresponds to the lower bound.
    
    The no-context baseline (upper bound on WER) corresponds to $\text{C}_t^{\text{none}} = \emptyset$
    meaning no contextual information is provided to the ASR system and corresponds to the upper bound.

    \section{Experimental Setup} \label{sec:setup}
    \subsection{Dataset}
    The experiments are conducted on three datasets: TED-LIUMv3 \cite{hernandez2018ted} (\textasciitilde 1.5 hours of audio), Earnings21 \cite{del2021earnings} (\textasciitilde 5 hours of audio) and SPGISpeech \cite{oneill2021spgispeech} (\textasciitilde 5 hours of audio). SPGISpeech consist of 5–15 seconds utterances grouped by sessions. Since context extraction methods require longer audio to capture contextual information, sessions were concatenated, and only segments longer than 6 minutes were retained, perseving topic consistency.
 
    To ensure consistency, transcripts are preprocessed before evaluation. Non-verbal content and non-English segments are removed. Text is normalized through lowercasing, punctuation stripping, hyphen removal, and conversion of numerical expressions to word format.
    
    For the vocabulary $\mathcal{V}$, a set of 466,358 unique, non-stop words\footnote{www.kaggle.com/datasets/bwandowando/479k-english-words \& NLTK Corpus Words (232k) www.nltk.org} is utilized for context generation using RAG. These words are combined together with their definitions (if available).
    
    Let $\mathcal{T}$ denote the set of entity types that are considered rare\footnote{\textit{Location, Organization, Geopolitical, Product, Person, Nationality-Religion-Political Groups}} and let $\mathcal{E_T}$ be the set of named entities of types in $\mathcal{T}$, extracted from the reference transcripts. We define the set of rare entities as $\mathcal{E}_{\text{rare}} = \{ e \in \mathcal{E_T} \mid  type(e) \in T \wedge e \notin \mathcal{S} \}$ and the set of out-of-vocabulary (OOV) entities as $\mathcal{E}_{\text{oov}} = \{ e \in \mathcal{E}_{\text{rare}} \mid e \notin \mathcal{V} \}$. To assess the lexical coverage and contextual demands of each dataset, we perform a rare entity analysis. Table~\ref{tab:rare_analysis} reports the percentage of unique words not present in the static vocabulary ($\mathcal{V}$) and the proportion of rare entities. 
    
    \begin{table}[H]
        \centering
        \vspace{0.2em}
        \resizebox{\columnwidth}{!}{
            \begin{tabular}{|l|c|c|c|} \hline
                \textbf{Metrics} & \textbf{TEDLIUMv3} & \textbf{Earnings21} & \textbf{SPGISpeech} \\ \hline
                OOV  & 6.31\% & 13.49\% & 15.77\%  \\ \hline
                Rare rate & 28\% & 38\% & 6.16\%  \\ \hline
            \end{tabular}
        }
        \caption{Out-of-vocabulary (OOV) and the percentage of rare words appearing across the speech datasets.}
        \label{tab:rare_analysis}
    \end{table}
    
    \begin{table*}[t]
        \fontsize{8.55pt}{8.55pt}\selectfont
        \setlength{\tabcolsep}{1pt}   
        \def\arraystretch{1.5}
        \begin{minipage}[t]{0.58\textwidth}
            \begin{tabular}{|l|cccc|cccc|cccc|}
                \hline
                {\textbf{Method [c, k]}} & \multicolumn{4}{c|}{\textbf{TED-LIUM v3}} & \multicolumn{4}{c|}{\textbf{Earnings21}} & \multicolumn{4}{c|}{\textbf{SPGISpeech}} \\
                & \textbf{WER $\downarrow$} & \textbf{Overlap $\uparrow$} & \textbf{Count $\downarrow$} & \textbf{Time $\downarrow$} 
                & \textbf{WER $\downarrow$} & \textbf{Overlap $\uparrow$} & \textbf{Count $\downarrow$} & \textbf{Time $\downarrow$} 
                & \textbf{WER $\downarrow$} & \textbf{Overlap $\uparrow$} & \textbf{Count $\downarrow$} & \textbf{Time $\downarrow$}\\
                \hline
                \verb|Oracle|            & 15.4\%          & 100\%           & 1$\times$              & --
                                         & 29.7\%          & 100\%           & 1$\times$              & --
                                         & 17.0\%          & 100\%           & 1$\times$              & --  \\
                \verb|No Context|        & 18.9\%          & --              & --                     & 1$\times$             
                                         & 35.9\%          & --              & --                     & 1$\times$    
                                         & 22.4\%          & --              & --                     & 1$\times$\\
                \verb|CB-LLM|            & 16.9\%          & 45.3\%          & 10.94$\times$          & 4.66$\times$
                                         & 31.8\%          & 52.9\%          & 9.03$\times$           & 3.26$\times$ 
                                         & \textbf{18.6\%} & 42.6\%          & 7.01$\times$           & 5.93$\times$\\
                \verb|CB-LLM LLM_fix|    & 16.8\%          & \textbf{48.7\%} & 10.84$\times$          & 6.16$\times$
                                         & 31.7\%          & \textbf{56.1\%} & 9.06$\times$           & 3.59$\times$ 
                                         & \textbf{18.6\%} & \textbf{43.7\%} & 6.3$\times$            & 6.56$\times$\\
                \verb|CB-RAG [100, 10]|  & 17.6\%          & 11\%            & 5.47$\times$           & 1.36$\times$          
                                         & 32.5\%          & 12.8\%          & 6.48$\times$           & 1.14$\times$ 
                                         & 18.7\%          & 8.8\%           & 3.41$\times$           & 1.4$\times$\\
                \verb|CB-RAG [100, 100]| & 17.6\%          & 6.3\%           & 2.67$\times$           & \textbf{1.02$\times$}
                                         & 31.6\%          & 8.8\%           & 3.81$\times$           & 1.13$\times$ 
                                         & 18.8\%          & 3.0\%           & 1.34$\times$           & \textbf{1.31$\times$}\\
                \verb|CB-RAG [250, 10]|  & \textbf{16.4\%} & 17.8\%          & \textbf{12$\times$}    & 1.16$\times$
                                         & \textbf{31.1\%} & 21.4\%          & \textbf{15.17$\times$} & 1.16$\times$ 
                                         & 18.7\%          & 14.5\%          & \textbf{9.40$\times$}  & 1.42$\times$\\
                \verb|CB-RAG [250, 100]| & 17.1\%          & 10.1\%          & \textbf{12$\times$}    & 1.04$\times$
                                         & 31.3\%          & 15.6\%          & 9.11$\times$           & \textbf{1.02$\times$} 
                                         & 18.8\%          & 5.5\%           & 2.3$\times$            & 1.34$\times$\\
                \hline
            \end{tabular}
        \end{minipage}
        \caption{Evaluation of context-extraction strategies on TED-LIUM v3, Earnings21 and SPGISpeech. Metrics include Word Error Rate (WER), Overlap percentage, Count of context words, and relative Time (normalized to no-context baseline).}
        \label{tab:context-results}
    \end{table*}
    \subsection{ASR Model}
    The context module of the ASR system implemented is based on the approach proposed by \cite{jalal2023locality}, which introduces a CB mechanism that integrates contextually relevant external information during inference. Following Figure \ref{fig:pipeline} we perform contextual biasing ASR per audio segment. For VAD, the SpeechBrain \cite{speechbrain_v1} library is utilized. After all segments are processed, their outputs are concatenated to reconstruct the full transcript.
    
    \subsection{Evaluation Metrics}
    The performance of the proposed contextual ASR system is evaluated using word error rate (WER), the standard metric for transcription accuracy. A contextual overlap score is calculated to assess how much of the ground-truth context is correctly recovered by each method, serving as a proxy for semantic recall. The size of the extracted context list per segment is also reported, indicating the expressive capacity of each method.

    \section{Results} \label{sec:results}
     Table~\ref{tab:context-results} presents the result of the proposed method and alternatives on TED-LIUMv3, Earnings21 and SPGISpeech. Regarding \verb|CB-RAG|, multiple configurations were investigated regarding the number of context words to retrieve with each query ($c$) and the number of segments to be used for the query construction ($k$).
    
    As shown in Table~\ref{tab:context-results}, the \verb|CB-RAG| method on TED-LIUMv3 exhibits a consistent reduction in WER as the number of retrieved contexts $c$ increases, from 17.6\% at $c$=100 to 16.4\% at $c$=250. Reducing the number of segments $k$ also results in improved performance: the [250,10] configuration yields a WER of 16.4\%, compared to 17.1\% for [250,100]. Although LLM-based methods achieve higher context overlap (45.3–48.7\%), \verb|CB-RAG| compensates through significantly higher context counts (up to 12\%), suggesting it access more diverse and redundant set of candidate segments. Regarding computational efficiency, \verb|CB-RAG| demonstrates substantially lower latency, 1.02–1.36 times slower than no-context baseline, relative to LLM-based approaches, which are 4.66–6.16 times. Among \verb|CB-LLM| variants, pre-correcting the ASR's transcript slightly improves WER (from 16.9\% to 16.8\%). Overall, the [250,10] \verb|CB-RAG| configuration provides the most balanced trade-off across accuracy, overlap, and latency, indicating strong suitability for real-time deployment.
    
    Similar trends are observed in the Earnings21 dataset. Increasing $c$ improves WER from 32.5\% at [100,10] to 31.1\% at [250, 10], approaching the LLM-based results ($\approx31.7\%$) despite lower overlap. LLM-based methods again show high overlap values, peaking at 56.1\%, while \verb|CB-RAG| ranges between 8.8\% and 21.4\%. The retrieval count for \verb|CB-RAG| is also considerably higher, with [250,10] reaching 15.17 compared to 9.06 for \verb|CB-LLM|. Latency for \verb|CB-RAG| remains much lower, ranging from 1.14 to 1.02, compared to \verb|CB-LLM| decoding times of 3.26 to 3.59. The [250,10] configuration again offers the best balance, with a WER of 31.1\%, overlap of 21.4\%, and a time cost of just 1.12. These results demonstrate \verb|CB-RAG|’s ability to scale effectively across different domains while maintaining a strong balance between accuracy and efficiency.
    
    Although LLM-based approaches achieve the best relative WER improvement in SPGISpeech (16.96\%), \verb|CB-RAG| configurations closely follow (16.52\%). As shown in Table~\ref{tab:rare_analysis}, the dataset features a high OOV rate and few rare entities. However, despite this lexical mismatch, \verb|CB-RAG| remains equivalent to the LLM-based methods. Reducing the number of segments $k$ from 100 to 10 again improves WER, reinforcing the value of recent focused context. As with other datasets, LLM-based methods show the highest contextual overlap (up to 43.7\%), while \verb|CB-RAG| achieves slightly higher context count (up to 9.4$\times$) and significantly lower latency, with the [100,100] configuration running at just 1.31$\times$ the cost of the \verb|No Context| baseline. These results underscore \verb|CB-RAG|’s efficiency and adaptability, even under challenging lexical conditions.
    
    The results indicate that the proposed \verb|CB-RAG| approach is a competitive and effective method for automatic context construction without the use of user-specific historical text data. It is also a better alternative compared to LLM-based context creation and transcript correction. Although \verb|CB-RAG| has lower contextual overlap scores, the WER is better with significantly lower latency, depicting its utility in actual scenarios. Its flexibility in selecting word context size extraction by each query ($c$) and number of segments in the past to consider ($k$) enables flexible application across domains with varied requirements for latency and performance. These findings suggest that \verb|CB-RAG| would be particularly appropriate for resource-constrained or real-time applications for which LLM-based alternatives would be computationally infeasible.
    
    \section{Conclusion} \label{sec:conlcusion}
    This paper explored different approaches to incorporating contextual information into ASR pipelines, with an emphasis on automatic context extraction. Among the approaches examined - embedding-based retrieval, context generated by LLM, and post-ASR correction - the proposed \verb|CB-RAG| framework achieved the best overall performance, yielding the lowest WER (up to a 17\% relative reduction) in test sets at the lowest computational cost and latency. Despite a substantially lower context overlap, up to 47.3\% absolute difference, \verb|CB-RAG| retrieved significantly higher number of contextual tokens, while operating, on average 83.5\%, lower latency than LLM-based alternatives. Post-ASR correction improved slightly but still remained less than that of \verb|CB-RAG|. These findings highlight \verb|CB-RAG| as the most scalable and adaptable solution, combining high accuracy with efficiency.
    
    \section*{Limitations}
    There are several limitations acknowledged. The \verb|CB-RAG| approach is lexically sensitive and might overlook semantically similar terms. It also assumes the presence of candidate context entries, which under unconstrained environments might not always be realistic. The \verb|CB-LLM| method depends on prompt quality and is prone to variability across model versions and decoding parameters. Also, the \verb|LLM-fix| is a post-processing step, and its performance tends to degrade in the presence of low-resource hardware or noisy transcription.
    
    \bibliography{custom}

\begin{thebibliography}{23}
\providecommand{\natexlab}[1]{#1}

\bibitem[{Baevski et~al.(2020)Baevski, Zhou, Mohamed, and
  Auli}]{Baevski2020wav2vec}
Alexei Baevski, Yuhao Zhou, Abdelrahman Mohamed, and Michael Auli. 2020.
\newblock {wav2vec 2.0}: A framework for self-supervised learning of speech
  representations.
\newblock In \emph{Advances in Neural Information Processing Systems
  (NeurIPS)}, volume~33, pages 12449--12460.

\bibitem[{Del~Rio et~al.(2021)Del~Rio, Delworth, Westerman, Huang, Bhandari,
  Palakapilly, McNamara, Dong, Zelasko, and Jett{\'e}}]{del2021earnings}
Miguel Del~Rio, Natalie Delworth, Ryan Westerman, Michelle Huang, Nishchal
  Bhandari, Joseph Palakapilly, Quinten McNamara, Joshua Dong, Piotr Zelasko,
  and Miguel Jett{\'e}. 2021.
\newblock Earnings-21: A practical benchmark for asr in the wild.
\newblock \emph{arXiv preprint arXiv:2104.11348}.

\bibitem[{Douze et~al.(2024)Douze, Guzhva, Deng, Johnson, Szilvasy, Mazaré,
  Lomeli, Hosseini, and Jégou}]{douze2024faiss}
Matthijs Douze, Alexandr Guzhva, Chengqi Deng, Jeff Johnson, Gergely Szilvasy,
  Pierre-Emmanuel Mazaré, Maria Lomeli, Lucas Hosseini, and Hervé Jégou.
  2024.
\newblock \href {https://arxiv.org/abs/2401.08281} {The faiss library}.

\bibitem[{Fu et~al.(2023)Fu, Sathyendra, Gandhe, Liu, Strimel, McGowan, and
  Mouchtaris}]{Fu2023Robust}
Xuandi Fu, Kanthashree~M. Sathyendra, Ankur Gandhe, Jing Liu, Grant~P. Strimel,
  Ross McGowan, and Athanasios Mouchtaris. 2023.
\newblock Robust acoustic and semantic contextual biasing in neural transducers
  for speech recognition.
\newblock In \emph{Proc. IEEE Intl. Conf. on Acoustics, Speech and Signal
  Processing (ICASSP)}, pages 1--5.

\bibitem[{Gourav et~al.(2021)Gourav, Liu, Gandhe, Gu, Lan, Huang, Kalmane,
  Tiwari, Filimonov, Rastrow, Stolcke, and Bulyko}]{Gourav2021}
Aditya Gourav, Linda Liu, Ankur Gandhe, Yile Gu, Guitang Lan, Xiangyang Huang,
  Shashank Kalmane, Gautam Tiwari, Denis Filimonov, Ariya Rastrow, Andreas
  Stolcke, and Ivan Bulyko. 2021.
\newblock Personalization strategies for end-to-end speech recognition systems.
\newblock In \emph{Proc. IEEE Int. Conf. Acoustics, Speech and Signal
  Processing (ICASSP)}, pages 7348--7352.

\bibitem[{Gulati et~al.(2020)Gulati, Qin, Chiu, Parmar, Zhang, Yu, Han, Wang,
  Zhang, Wu, and Pang}]{Gulati2020Conformer}
Anmol Gulati, James Qin, Chung-Cheng Chiu, Niki Parmar, Yu~Zhang, Jiahui Yu,
  Wei Han, Shibo Wang, Zhengdong Zhang, Yonghui Wu, and Ruoming Pang. 2020.
\newblock Conformer: Convolution-augmented transformer for speech recognition.
\newblock In \emph{Proc. Interspeech}, pages 5036--5040.

\bibitem[{Han et~al.(2022)Han, Dong, Liang, Cai, Zhou, Ma, and
  Xu}]{Han2022FineCoS}
Minglun Han, Linhao Dong, Zhenlin Liang, Meng Cai, Shiyu Zhou, Zejun Ma, and
  Bo~Xu. 2022.
\newblock Improving end-to-end contextual speech recognition with fine-grained
  contextual knowledge selection.
\newblock In \emph{Proc. IEEE Intl. Conf. on Acoustics, Speech and Signal
  Processing (ICASSP)}, pages 8532--8536.

\bibitem[{Hernandez et~al.(2018)Hernandez, Nguyen, Ghannay, Tomashenko, and
  Esteve}]{hernandez2018ted}
Fran{\c{c}}ois Hernandez, Vincent Nguyen, Sahar Ghannay, Natalia Tomashenko,
  and Yannick Esteve. 2018.
\newblock Ted-lium 3: Twice as much data and corpus repartition for experiments
  on speaker adaptation.
\newblock In \emph{Speech and Computer: 20th International Conference, SPECOM
  2018, Leipzig, Germany, September 18--22, 2018, Proceedings 20}, pages
  198--208. Springer.

\bibitem[{Huang et~al.(2024)Huang, Yarmohammadi, Khudanpur, and
  Povey}]{huang2024improving}
Ruizhe Huang, Mahsa Yarmohammadi, Sanjeev Khudanpur, and Daniel Povey. 2024.
\newblock Improving neural biasing for contextual speech recognition by early
  context injection and text perturbation.
\newblock \emph{arXiv preprint arXiv:2407.10303}.

\bibitem[{Jain et~al.(2020)Jain, Keren, Mahadeokar, Zweig, Metze, and
  Saraf}]{Jain2020Contextual}
Mahaveer Jain, Gil Keren, Jay Mahadeokar, Geoffrey Zweig, Florian Metze, and
  Yatharth Saraf. 2020.
\newblock Contextual rnn-t for open domain {ASR}.
\newblock In \emph{Proc. Interspeech}, pages 11--15.

\bibitem[{Jalal et~al.(2023)Jalal, Parada, Pavlidis, Moschopoulos, Saravanan,
  Kontoulis, Zhang, Drosou, Lee, Lee et~al.}]{jalal2023locality}
Md~Asif Jalal, Pablo~Peso Parada, George Pavlidis, Vasileios Moschopoulos,
  Karthikeyan Saravanan, Chrysovalantis-Giorgos Kontoulis, Jisi Zhang,
  Anastasios Drosou, Gil~Ho Lee, Jungin Lee, Seokyeong Jung. 2023.
\newblock Locality enhanced dynamic biasing and sampling strategies for
  contextual asr.
\newblock In \emph{2023 IEEE Automatic Speech Recognition and Understanding
  Workshop (ASRU)}, pages 1--8. IEEE.

\bibitem[{Mathur et~al.(2024)Mathur, Liu, Li, Ma, Karen, Ahmed, Manocha, and
  Zhang}]{mathur-etal-2024-doc}
Puneet Mathur, Zhe Liu, Ke~Li, Yingyi Ma, Gil Karen, Zeeshan Ahmed, Dinesh
  Manocha, and Xuedong Zhang. 2024.
\newblock \href {https://aclanthology.org/2024.lrec-main.457/} {{DOC}-{RAG}:
  {ASR} language model personalization with domain-distributed co-occurrence
  retrieval augmentation}.
\newblock In \emph{Proceedings of the 2024 Joint International Conference on
  Computational Linguistics, Language Resources and Evaluation (LREC-COLING
  2024)}, pages 5132--5139, Torino, Italia. ELRA and ICCL.

\bibitem[{Meng et~al.(2024)Meng, Wu, Prabhavalkar, Peyser, Wang, Chen, Sainath,
  and Ramabhadran}]{meng2024text}
Zhong Meng, Zelin Wu, Rohit Prabhavalkar, Cal Peyser, Weiran Wang, Nanxin Chen,
  Tara~N Sainath, and Bhuvana Ramabhadran. 2024.
\newblock Text injection for neural contextual biasing.
\newblock \emph{arXiv preprint arXiv:2406.02921}.

\bibitem[{Muralidhar~Jayanthi et~al.(2023)Muralidhar~Jayanthi, Kulshreshtha,
  Dingliwal, Ronanki, and Bodapati}]{muralidhar2023retrieve}
Sai Muralidhar~Jayanthi, Devang Kulshreshtha, Saket Dingliwal, Srikanth
  Ronanki, and Sravan Bodapati. 2023.
\newblock Retrieve and copy: Scaling asr personalization to large catalogs.
\newblock \emph{arXiv e-prints}, pages arXiv--2311.

\bibitem[{Park et~al.(2019)Park, Chan, Zhang, Chiu, Zoph, Cubuk, and
  Le}]{Park2019SpecAugment}
Daniel~S. Park, William Chan, Yu~Zhang, Chung-Cheng Chiu, Barret Zoph, Ekin~D.
  Cubuk, and Quoc~V. Le. 2019.
\newblock {SpecAugment}: A simple data augmentation method for automatic speech
  recognition.
\newblock In \emph{Proc. Interspeech}, pages 2613--2617.

\bibitem[{Ravanelli et~al.(2024)Ravanelli, Parcollet, Moumen, de~Langen,
  Subakan, Plantinga, Wang, Mousavi, Libera, Ploujnikov, Paissan, Borra, Zaiem,
  Zhao, Zhang, Karakasidis, Yeh, Champion, Rouhe, Braun, Mai, Zuluaga-Gomez,
  Mousavi, Nautsch, Nguyen, Liu, Sagar, Duret, Mdhaffar, Laperri{{\`e}}re,
  Rouvier, Mori, and Est{{\`e}}ve}]{speechbrain_v1}
Mirco Ravanelli, Titouan Parcollet, Adel Moumen, Sylvain de~Langen, Cem
  Subakan, Peter Plantinga, Yingzhi Wang, Pooneh Mousavi, Luca~Della Libera,
  Artem Ploujnikov, Francesco Paissan, Davide Borra, Salah Zaiem, Zeyu Zhao,
  Shucong Zhang, Georgios Karakasidis, Sung-Lin Yeh, Pierre Champion, Aku
  Rouhe, and 14 others. 2024.
\newblock \href {http://jmlr.org/papers/v25/24-0991.html} {Open-source
  conversational ai with speechbrain 1.0}.
\newblock \emph{Journal of Machine Learning Research}, 25(333).

\bibitem[{Sainath et~al.(2023)Sainath, Prabhavalkar, Caseiro, Rondon, and
  Allauzen}]{Sainath2023}
Tara~N. Sainath, Rohit Prabhavalkar, Diamantino Caseiro, Patrick Rondon, and
  Cyril Allauzen. 2023.
\newblock Improving contextual biasing with text injection.
\newblock In \emph{Proc. IEEE Int. Conf. Acoustics, Speech and Signal
  Processing (ICASSP)}, pages 1--5.

\bibitem[{Sathyendra et~al.(2022)Sathyendra, Muniyappa, Chang, Liu, Su,
  Strimel, Mouchtaris, and Kunzmann}]{MysoreSathyendra2022}
Kanthashree~Mysore Sathyendra, Thejaswi Muniyappa, Feng{-}Ju Chang, Jing Liu,
  Jinru Su, Grant~P. Strimel, Athanasios Mouchtaris, and Siegfried Kunzmann.
  2022.
\newblock Contextual adapters for personalized speech recognition in neural
  transducers.
\newblock In \emph{Proc. IEEE Int. Conf. Acoustics, Speech and Signal
  Processing (ICASSP)}, pages 8537--8541.

\bibitem[{Sun et~al.(2024)Sun, Ahmed, Ma, Liu, Kabela, Pang, and
  Kalinli}]{sun2024contextual}
Chuanneng Sun, Zeeshan Ahmed, Yingyi Ma, Zhe Liu, Lucas Kabela, Yutong Pang,
  and Ozlem Kalinli. 2024.
\newblock Contextual biasing of named-entities with large language models.
\newblock In \emph{ICASSP 2024-2024 IEEE International Conference on Acoustics,
  Speech and Signal Processing (ICASSP)}, pages 10151--10155. IEEE.

\bibitem[{Tong et~al.(2023{\natexlab{a}})Tong, Harding, and
  Wiesler}]{tong2023hierarchical}
Sibo Tong, Philip Harding, and Simon Wiesler. 2023{\natexlab{a}}.
\newblock Hierarchical attention-based contextual biasing for personalized
  speech recognition using neural transducers.
\newblock In \emph{2023 IEEE Automatic Speech Recognition and Understanding
  Workshop (ASRU)}, pages 1--8. IEEE.

\bibitem[{Tong et~al.(2023{\natexlab{b}})Tong, Harding, and
  Wiesler}]{tong2023slot}
Sibo Tong, Philip Harding, and Simon Wiesler. 2023{\natexlab{b}}.
\newblock Slot-triggered contextual biasing for personalized speech recognition
  using neural transducers.
\newblock In \emph{ICASSP 2023-2023 IEEE International Conference on Acoustics,
  Speech and Signal Processing (ICASSP)}, pages 1--5. IEEE.

\bibitem[{Wu et~al.(2024)Wu, Song, Li, Rondon, Meng, Velez, Wang, Caseiro,
  Pundak, Munkhdalai, Chandorkar, and Prabhavalkar}]{Wu2024}
Zelin Wu, Gan Song, Christopher Li, Patrick Rondon, Zhong Meng, Xavier Velez,
  Weiran Wang, Diamantino Caseiro, Golan Pundak, Tsendsuren Munkhdalai, Angad
  Chandorkar, and Rohit Prabhavalkar. 2024.
\newblock {Deferred NAM}: Low-latency top-k context injection via deferred
  context encoding for non-streaming {ASR}.
\newblock In \emph{Proc. NAACL-HLT (Industry Track)}, pages 315--323.

\bibitem[{Xiao et~al.(2025)Xiao, Hou, Garcia-Romero, and
  Han}]{xiao2025contextual}
Cihan Xiao, Zejiang Hou, Daniel Garcia-Romero, and Kyu~J Han. 2025.
\newblock Contextual asr with retrieval augmented large language model.
\newblock In \emph{ICASSP 2025-2025 IEEE International Conference on Acoustics,
  Speech and Signal Processing (ICASSP)}, pages 1--5. IEEE.

\end{thebibliography}
    
    \clearpage
    \section*{Appendix}
    \label{sec:appendix}
    \subsection*{Prompt for LLM-driven Context Generation}
    \emph{"You are the master of knowledge, with expertise in every domain. Given a sentence and based on your knowledge, provide a huge number of relevant words. Focus on names, locations, terminology, concepts. Provide only the words, comma-separated, without any other explanations."}
    \subsection*{Prompt for post-ASR Transcript Refinement using LLM}
    \emph{"You are a master philologist and grammar expert. Using the provided conversation history for context, correct the given sentence by fixing typos, misspellings, grammar, or logical inconsistencies. Preserve the original intent. Respond with only the revised sentence, nothing else."}
    \subsection*{Zipf Distribution of Word Frequencies} \label{app:zipf}
    \begin{figure}[ht]
        \centering
        \includegraphics[width=\linewidth]{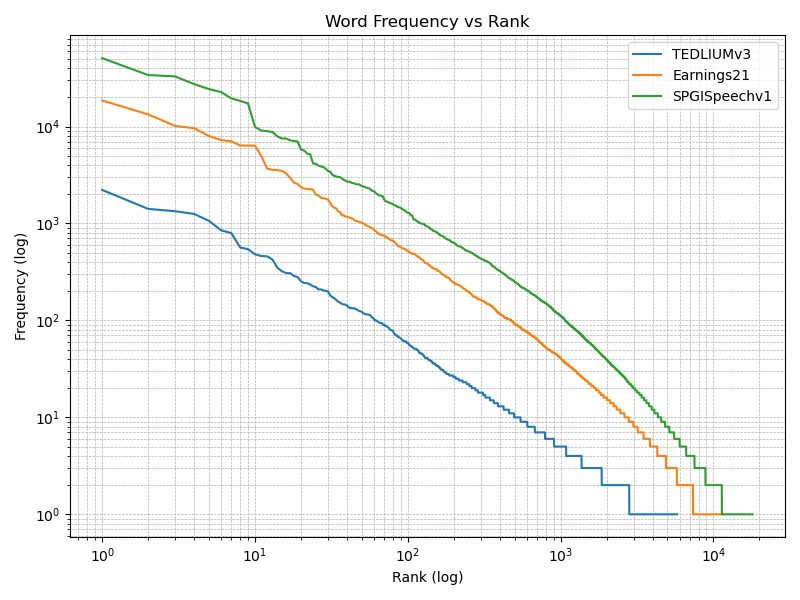}  
        \caption{Zipf distribution of word frequencies~\cite{piantadosi2014zipf} across all datasets, confirming the characteristic long-tail structure of natural language.}
        \label{fig:zipf_plot}
    \end{figure}
    
    \begin{table*}[t]
        \fontsize{9pt}{9pt}\selectfont
        \setlength{\tabcolsep}{1pt} 
        \def\arraystretch{1.3}
        \begin{minipage}[t]{\textwidth}
            \begin{tabular}{|l@{\hspace{6pt}}|l@{\hspace{7pt}}|cccc|cccc|}
                \hline
                {\textbf{Method}} & {\textbf{Model}} & \multicolumn{4}{c|}{\textbf{TED-LIUM v3}} & \multicolumn{4}{c|}{\textbf{Earnings21}} \\
                & & \textbf{WER $\downarrow$} & \textbf{Overlap $\uparrow$} & \textbf{Count $\downarrow$} & \textbf{Time $\downarrow$} 
                  & \textbf{WER $\downarrow$} & \textbf{Overlap $\uparrow$} & \textbf{Count $\downarrow$} & \textbf{Time $\downarrow$} \\
                \hline
                \verb|Oracle| & \multicolumn{1}{c|}{--} & 15.4\%   & 100\%        & 1$\times$    & -- & 29.7\%   & 100\%         & 1$\times$   & --         \\
                \verb|No Context| & \multicolumn{1}{c|}{--} & 18.9\%   & --           & --           & 1$\times$ & 35.9\%   & --            & --          & 1$\times$\\
                \hline
                \multirow{7}{*}{\texttt{CB-LLM}}
                    & \verb|llama3.2|       & 16.9\% & 45.3\%          & 10.94$\times$          & 4.66$\times$           
                                            & 31.8\% & 52.9\%          & 9.03$\times$           & 3.26$\times$ \\
                    & \verb|olmo2:7b|       & 16.8\% & 60.4\%          & 8.65$\times$           & 9.43$\times$           
                                            & 32.5\% & 65.3\%          & 7.58$\times$           & 8.33$\times$ \\
                    & \verb|gemma3:4b|      & 16.9\% & 58.4\%          & 13$\times$             & 21.86$\times$ 
                                            & 32.5\% & 66.7\%          & 8.74$\times$           & 28.95$\times$ \\
                    & \verb|tinyllama:1.1b| & 16.9\% & \textbf{81.8\%} &\textbf{19.73$\times$}  & 9.61$\times$           
                                            & 32.4\% & 76.5\%          & 10.92$\times$          & 6.94$\times$ \\
                    & \verb|smollm2:135m|   & 16.6\% & 77.1\%          & 6.1$\times$            & 4.01$\times$           
                                            & 32.8\% & \textbf{79.1\%} & 5.58$\times$           & 5.48$\times$ \\
                    & \verb|smollm360m|     & 16.7\% & 66.1\%          & 6.1$\times$            & 3.77$\times$           
                                            & 33.1\% & 73.9\%          & 5.33$\times$           & 5.62$\times$ \\
                    & \verb|qwen2.5:0.5b|   & 16.8\% & 55.6\%          & 7.52$\times$           & 5.21$\times$           
                                            & 32.6\% & 57.0\%          & 6.16$\times$           & 6.03$\times$ \\
                \hline
                \multirow{7}{*}{\texttt{CB-LLM \& LLM\_fix}}
                    & \verb|llama3.2|       & 16.8\% & 48.7\% & 10.84$\times$ & 6.16$\times$  
                                            & 31.7\% & 56.1\% & 9.06$\times$  & 3.59$\times$ \\
                    & \verb|olmo2:7b|       & 16.8\% & 57.6\% & 9.18$\times$  & 17.87$\times$ 
                                            & 32.7\% & 65.7\% & 7.53$\times$  & 15.47$\times$ \\
                    & \verb|gemma3:4b|      & 16.9\% & 57.6\% & 12.77$\times$ & 25.06$\times$ 
                                            & 32.5\% & 63.0\% & 8.44$\times$  & 33.58$\times$ \\
                    & \verb|tinyllama:1.1b| & 16.8\% & 80.1\% & 18.85$\times$ & 10.01$\times$ 
                                            & 32.9\% & 77.7\% & 10.24$\times$ & 7.12$\times$ \\
                    & \verb|smollm2:135m|   & 16.7\% & 65.9\% & 6.08$\times$  & 4.52$\times$ 
                                            & 33.1\% & 71.1\% & 4.64$\times$  & 6.63$\times$ \\
                    & \verb|smollm360m|     & 16.8\% & 64.8\% & 5.31$\times$  & 4.98$\times$  
                                            & 33.0\% & 71.5\% & 4.46$\times$  & 7.06$\times$ \\
                    & \verb|qwen2.5:0.5b|   & 16.8\% & 50.2\% & 6.81$\times$  & 5.82$\times$  
                                            & 33.0\% & 55.5\% & 5.88$\times$  & 7.41$\times$ \\
                \hline
                \multirow{4}{*}{\texttt{CB-RAG [100, 10]}}
                    & \verb|all-MiniLM-L6-v2|     & 17.6\% & 11\%   & 5.47$\times$ & 1.36$\times$ 
                                                  & 32.5\% & 12.8\% & 6.48$\times$ & 1.14$\times$ \\
                    & \verb|all-MiniLM-L12-v2|    & 17.0\% & 13.1\% & 7.64$\times$ & 1.08$\times$ 
                                                  & 34.5\% & 16.7\% & 9.98$\times$ & 1.08$\times$ \\
                    & \verb|all-distilroberta-v1| & 17.0\% & 9.6\%  & 5.63$\times$ & 1.30$\times$ 
                                                  & 35.0\% & 11.5\% & 5.77$\times$ & 1.20$\times$ \\
                    & \verb|all-mpnet-base-v2|    & 17.0\% & 10.5\% & 6.21$\times$ & 1.31$\times$  
                                                  & 35.0\% & 9.1\%  & 6.00$\times$ & 1.10$\times$ \\
                \hline
                \multirow{4}{*}{\texttt{CB-RAG [100, 100]}}
                    & \verb|all-MiniLM-L6-v2|     & 17.6\% & 6.3\%  & 2.67$\times$ & 1.02$\times$ 
                                                  & 31.6\% & 8.8\%  & 3.81$\times$ & 1.13$\times$ \\
                    & \verb|all-MiniLM-L12-v2|    & 17.1\% & 8.3\%  & 4.05$\times$ & 1.01$\times$ 
                                                  & 35.2\% & 14.6\% & 8.37$\times$ & 1.10$\times$ \\
                    & \verb|all-distilroberta-v1| & 17.0\% & 4.4\%  & 2.16$\times$ & 1.30$\times$ 
                                                  & 35.0\% & 6.4\%  & 2.76$\times$ & 1.16$\times$ \\
                    & \verb|all-mpnet-base-v2|    & 17.0\% & 3.9\%  & 2.01$\times$ & 1.27$\times$ 
                                                  & 35.1\% & 3.3\%  & 2.07$\times$ & 1.19$\times$ \\    
                \hline
                \multirow{4}{*}{\texttt{CB-RAG [250, 10]}}
                    & \verb|all-MiniLM-L6-v2|     & \textbf{16.4\%} & 17.8\% & 12$\times$             & 1.16$\times$ 
                                                  & \textbf{31.1\%} & 21.4\% & \textbf{15.17$\times$} & 1.16$\times$ \\
                    & \verb|all-MiniLM-L12-v2|    & 17.0\%          & 20.4\% & 17.6$\times$           & 1.14$\times$ 
                                                  & 34.4\%          & 18.7\% & 13.82$\times$          & 1.20$\times$ \\
                    & \verb|all-distilroberta-v1| & 17.0\%          & 15.2\% & 12.94$\times$          & 1.15$\times$ 
                                                  & 34.8\%          & 18.7\% & 11.89$\times$          & 1.24$\times$ \\
                    & \verb|all-mpnet-base-v2|    & 17.1\%          & 16.6\% & 14.48$\times$          & 1.15$\times$ 
                                                  & 35.0\%          & 15.1\% & 12.05$\times$          & 1.22$\times$ \\  
                \hline
                \multirow{4}{*}{\texttt{CB-RAG [250, 100]}}
                    & \verb|all-MiniLM-L6-v2|     & 17.1\%          & 10.1\% & 12$\times$             & 1.04$\times$ 
                                                  & 31.3\%          & 15.6\% & 9.11$\times$           & \textbf{1.02$\times$} \\
                    & \verb|all-MiniLM-L12-v2|    & 17.0\%          & 13\%   & 9.69$\times$           & \textbf{1.01$\times$}
                                                  & 34.6\%          & 17.6\% & 12.46$\times$          & 1.05$\times$ \\
                    & \verb|all-distilroberta-v1| & 17.0\%          & 7.4\%  & 5.41$\times$           & 1.03$\times$ 
                                                  & 35.1\%          & 11.6\% & 6.57$\times$           & 1.14$\times$ \\
                    & \verb|all-mpnet-base-v2|    & 17.0\%          & 6.9\%  & 5.3$\times$            & 1.03$\times$ 
                                                  & 35.1\%          & 6.8\%  & 5.3$\times$            & 1.15$\times$ \\           
                \hline
            \end{tabular}
        \end{minipage}
        \caption{Evaluation of context-extraction strategies on TED-LIUM v3 and Earnings21 for various LLM models and different sentence transformers. Metrics include Word Error Rate (WER), Overlap percentage, Count of context words, and relative Time (normalized to no-context baseline).}
        \label{tab:context-results-additional}
    \end{table*}
    
    \subsection*{Impact of Model and Encoder Choice}
    Table~\ref{tab:context-results-additional} provides a deeper comparison of different model backbones used within the \verb|CB-LLM|, \verb|CB-LLM & LLM_fix| and \verb|CB-RAG| frameworks across TED-LIUMv3 and Earnings21.
    
    In TED-LIUMv3, the best WER (16.4\%) is achieved by \verb|CB-RAG [250,10]| using \verb|all-MiniLM-L6-v2|, representing a 13.2\% relative improvement over the no-context baseline.  
    Among LLM-based methods, \verb|smollm2:135m| delivers the strongest performance with a WER of 16.6\%, followed closely by \verb|smollm2:365m|, which is the faster among LLM variants. The highest contextual overlap (81.8\%) and context tokens (19.73$\times$) are achieved by the \verb|tinyllama:1.1b|, while \verb|gemma2:4b| is significantly the slowest. The \verb|LLM_fix| mechanism introduces a mean latency increase of approximately 2.3 times, affecting larger models like \verb|olmo2:7b| the most, while impacting smaller ones like \verb|tinyllama| much less. 
    
    Similar trends are observed in Earnings21, where the best WER (31.1\%) is again obtained by \verb|CB-RAG [250,10]| with \verb|all-MiniLM-L6-v2|—corresponding to a 13.4\% relative improvement. This configuration also yields the highest retrieval count (15.17 times than \verb|Oracle|), while \verb|smollm2:135m| achieves the highest contextual overlap (79.1\%) among LLM-based approaches. As with TED-LIUMv3, LLM-based correction is slowing down transcription by approximately 2.3 times, with \verb|olmo2:7b| experiencing the steepest latency increase (around 7 times). Interestingly, \verb|llama3.2| runs faster than both large and mini LLM models. Among LLMs, \verb|tinyllama| again generates the most context tokens, while \verb|smollm2| leads in overlap. Again, \verb|CB-RAG| encoders based on \verb|MiniLM| demonstrate high efficiency, with the [250,100] configuration achieving both high accuracy and the fastest runtime across all models (approximately 1.02 times).
    
    Across both datasets, \verb|CB-RAG| with \verb|all-MiniLM-L6-v2| stands out as the most effective model, offering the best WERs, high retrieval counts, and low latency. The [250,10] configuration consistently perform well, balancing accuracy and speed. Among LLMs, \verb|smollm2:135m| surpasses larger models such as \verb|llama3.2| in both WER and efficiency, making it a strong choice for constrained environments. Within the \verb|CB-RAG| framework, \verb|MiniLM-L6-v2| also achieves the highest contextual overlap (18.1\% more than the lowest-performing \verb|mpnet-base-v2|). The fastest encoder is \verb|MiniLM-L12-v2|, outperforming the slowest models (\verb|distilroberta| and \verb|mpnet-base-v2|) by approximately 7\% in runtime. These results reinforce the importance of model selection in both LLM-based and retrieval-based pipelines and highlight that compact models, when properly configured, can rival or even outperform larger architectures.
    
    The choice of backbone models plays a critical role in balancing quality and efficiency. Within \verb|CB-RAG|, the \verb|all-MiniLM-L6-v2| encoder offered the best trade-off, consistently delivering the highest accuracy and fastest inference. For LLM-based methods, \verb|llama3.2| achieves competitive performance and speed, however, in scenarios with stricter latency constraints, smaller models like \verb|smollm2:135m| are necessary to maintain responsiveness. 
    
\end{document}